\documentclass{article}
\usepackage{ijcai25}

\usepackage{times}
\usepackage{soul}
\usepackage{url}
\usepackage[hidelinks]{hyperref}
\usepackage[utf8]{inputenc}
\usepackage[small]{caption}
\usepackage{graphicx}
\usepackage{amsmath}
\usepackage{amsthm}
\usepackage{amsfonts}
\usepackage{booktabs}
\usepackage{algorithm}
\usepackage{algorithmic}
\usepackage{booktabs}
\usepackage{tabularx}
\usepackage{multirow}
\usepackage[switch]{lineno}

\title{RouteWinFormer: A Route-Window Transformer for Middle-range Attention in Image Restoration}

\author{
Qifan Li$^1$
\and
Tianyi Liang$^1$\and
Xingtao Wang$^1$\And
Xiaopeng Fan*$^1$\\
\affiliations
$^1$Faculty of Computing, Harbin Institute of Technology, Harbin 150001, China\\
\emails
\{23b903056,23S103169\}@stu.hit.edu.cn,\{xtwang,fxp\}@hit.edu.cn}

\begin{document}

\maketitle

\begin{abstract}
Transformer models have recently garnered significant attention in image restoration due to their ability to capture long-range pixel dependencies. However, long-range attention often results in computational overhead without practical necessity, as degradation and context are typically localized. Normalized average attention distance across various degradation datasets shows that middle-range attention is enough for image restoration. Building on this insight, we propose RouteWinFormer, a novel window-based Transformer that models middle-range context for image restoration. RouteWinFormer incorporates Route-Windows Attnetion Module, which dynamically selects relevant nearby windows based on regional similarity for attention aggregation, extending the receptive field to a mid-range size efficiently. In addition, we introduce Multi-Scale Structure Regularization during training, enabling the sub-scale of the U-shaped network to focus on structural information, while the original-scale learns degradation patterns based on generalized image structure priors. Extensive experiments demonstrate that RouteWinFormer outperforms state-of-the-art methods across 9 datasets in various image restoration tasks.

\end{abstract}


\section{Introduction}

The degradation (like blur, snowflake, haze, and raindrop) in images destroys critical contextual information, resulting in performance decline in subsequent tasks such as autonomous driving \cite{ref1}, medical imaging \cite{ref2}, remote sensing \cite{ref3}, and etc. Image restoration, which aims to produce an enhanced image from its degraded version, typically serves as a preliminary step for numerous practical applications. Given the ill-posed nature of image restoration, traditional image restoration approaches \cite{dcp,ref4} rely on carefully handcrafted prior assumptions to restrict the solution space to natural images. However, these traditional methods often exhibit high complexity and are insufficient for handling complex real-world scenarios \cite{mprnet}.

\begin{figure}[!t]
\centering
\includegraphics[width=\linewidth]{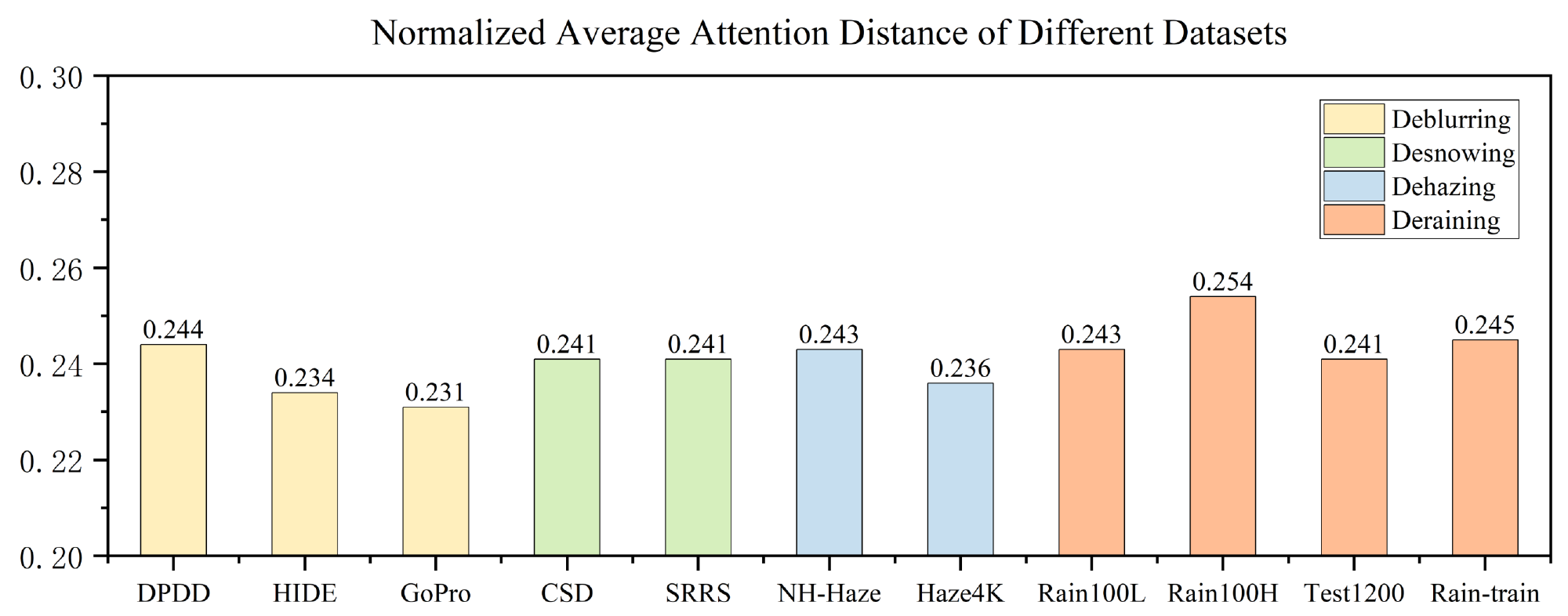}
\caption{Normalized average attention distance across various degradation datasets, where attention intensity is weighted by pixel distance and normalized by image size. A larger value indicates a broader attention area. All datasets show values below 0.3, suggesting that long-range modeling may not be practical necessity for image restoration.}
\label{motivate}
\end{figure}

With the rapid advancements in deep learning, Convolutional Neural Networks (CNNs) and Transformer models have achieved remarkable success, surpassing traditional methods in image restoration. CNNs \cite{naf} employ convolution operations to implicitly capture local details at the pixel level. However, their limited receptive fields and the static nature of convolutional filters restrict their flexibility in generalizing to varying input content \cite{restormer}. Directly applying naive Transformers \cite{transformer} to image restoration leads to high computational and memory complexity. Consequently, recent research has focused on developing more efficient structures for modeling long-range attention, such as transposed attention \cite{restormer} and anchored stripe attention \cite{grl}. Despite these advancements, these methods still incur considerable computational overhead. Alternatively, some approaches \cite{swinIR,cat} have sought to reduce computational and memory costs by restricting the attention field. However, these methods typically confine attention modeling to rectangular windows with shift operations, which are limited to capturing fixed local attention. Therefore, achieving a balance between capturing long-range dependencies and reducing computational overhead remains a critical challenge in image restoration research.

Image degradation can be broadly divided into local and global degradation. Local degradations (e.g.,defocus blur, snowflake) typically occur in specific regions and primarily impact nearby areas, while global degradations (e.g., haze, low light) rely on contextual information from surrounding regions for effective restoration. This raises the question: is it necessary to model long-range attention across the entire image for image restoration? We compute the normalized average attention distance \cite{ViT} across various degradation datasets, as shown in Fig. \ref{motivate}. This distance correlates with the receptive field size, where a larger value indicates a broader attention span. The results reveal that the normalized average attention distance across all degradation datasets remains below 0.3, indicating that long-range attention is not a practical necessity for image restoration. 

Building on this insight, we propose a novel window-based Transformer, called RouteWinFormer, to efficiently capture middle-range attention for image restoration. RouteWinFormer enables pixel-level context interaction across windows and is not constrained by fixed window sizes. It incorporates Route-windows Attention Module that dynamically selects the most relevant nearby windows based on regional similarity for attention aggregation, thereby extending the receptive field to a mid-range distance in a computationally efficient manner. In addition, degradation and context patterns often exhibit stripe-like or rectangle-like structures. To capture these patterns, we introduce two types of regional similarity: cross-shaped and rectangle-shaped regions. Besides, we introduce a Multi-Scale Structure Regularization (MSR) term into the loss function, which leverages structure information in clean image at the sub-scales to generalize the learning process. MSR encourages the sub-scales to focus on learning the textures and structures of the image, providing crucial priors that enable the original-scale to better generalize the learned degradation patterns. Notably, MSR can be naturally integrated into a multi-scale network.



The main contributions of this paper are as follows:
\begin{itemize}
\item[$\bullet$]
We present RouteWinFormer, a novel window-based Transformer designed to efficiently capture middle-range attention for image restoration. It incorporates Route-windows Attention Module that dynamically selects the most relevant nearby windows for attention aggregation, utilizing two types of regional similarity: cross-shaped and rectangle-shaped regions.

\item[$\bullet$]
We introduce Multi-Scale Structure Regularization, which directs the sub-scales to focus on capturing the textures and structures of the image, providing crucial priors that enhance the generalization of learned degradation patterns, thereby improving image restoration.

\item[$\bullet$] 
Extensive experiments validate that our model achieves state-of-the-art performance across 9 benchmark datasets spanning four representative image restoration tasks.

\end{itemize}

\begin{figure*}[!t]
\centering
\includegraphics[scale = 0.45]{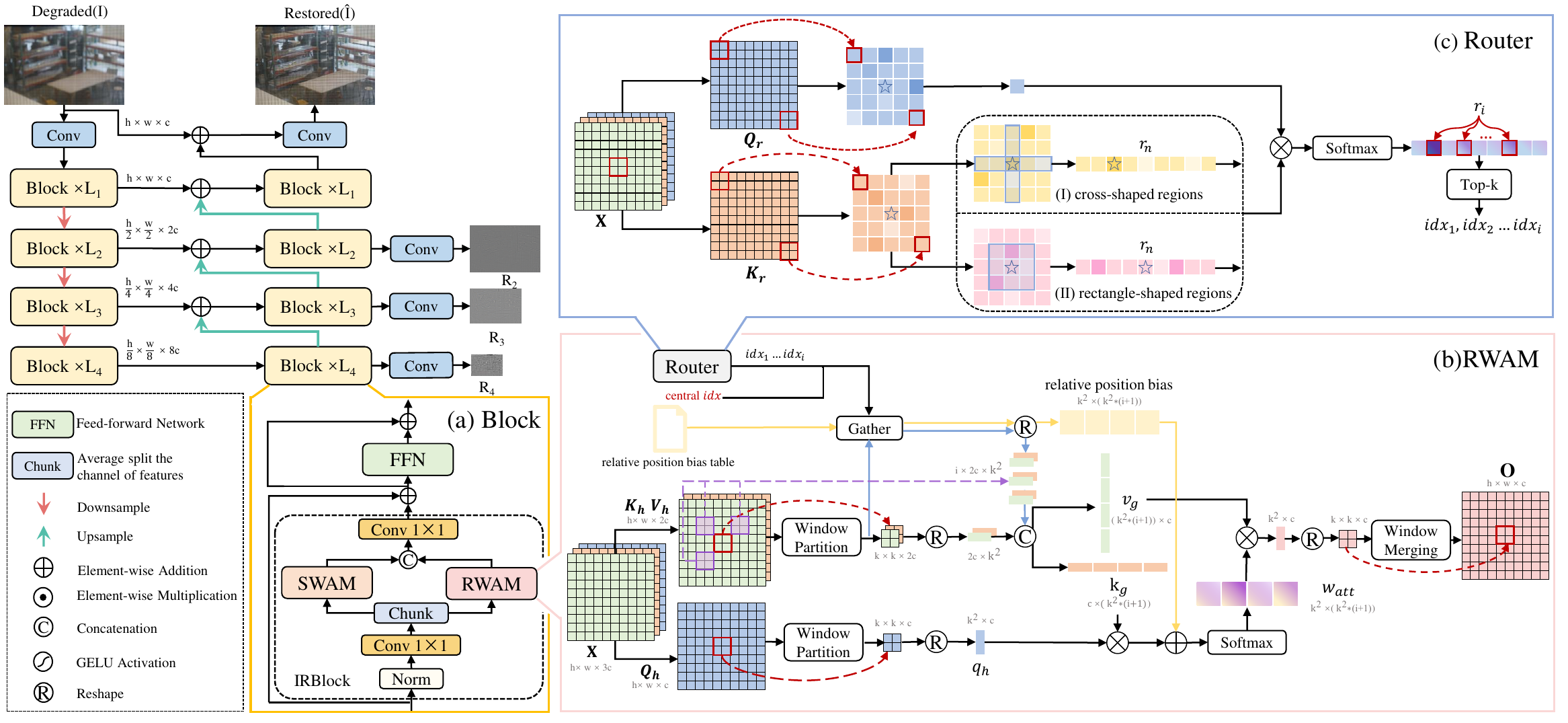}
\caption{The architecture of RouteWinFormer: (a) Transformer Block with Image Restoration Block (IRBlock) and Feed-Forward Network (FFN), incorporating Route-Windows Attention Module (RWAM) and Shift-Windows Attention Module (SWAM). (b) RWAM for aggregating middle-range contextual information. (c) In RWAM, the Router dynamically selects relevant nearby windows for attention aggregation based on regional similarities.}
\label{arch}
\end{figure*}

\section{Related Work}
\subsection{Image Restoration}

Image restoration aims to reconstruct an enhanced image from its degraded counterpart, enhancing visual experience and providing high-quality inputs for downstream high-level vision tasks. Following the development of deep learning method, CNNs have surpassed traditional methods \cite{traditional_1,traditional_2,traditional_3} as the dominant technique for image restoration, achieving remarkable performance across various subtasks such as defocus deblurring \cite{defocus1,deblurring_2,dpdd}, deraining \cite{deraining_1,deraining_2,deraining_3}, desnowing \cite{desnownet,csd,srrs} and dehazing \cite{dehazing_1,dehazing_2,dehazing_3}. The Transformer has demonstrated a powerful ability to model long sequences in natural language processing tasks \cite{transformer}. However, directly applying long-range attention to capture degradation or context patterns may provide some benefits, it incurs significant computational overhead and memory costs that outweigh its practical advantages.

\subsection{Vision Transformers}
Building on the success of Transformer architectures \cite{transformer,swin} in high-level computer vision tasks, researchers have introduced Transformer models into low-level vision tasks to capture attention between pixels. However, the quadratic complexity associated with input dimensions has become a bottleneck for vision transformers. In order to reduce the computational complexity of self-attention, some approaches aim to simplify calculations across the global scope. \cite{restormer} introduce transposed attention by exchanging spatial and channel dimensions. \cite{grl} conduct anchored self-attention within vertical and horizontal stripes, enabling a global view of the image content with reduced computational complexity. Others focus on limiting the scope of self-attention and model attention within local windows. \cite{swinIR} builds upon the Swin Transformer architecture \cite{swin}, leveraging window-based self-attention and shifted window operations to achieve outstanding image restoration performance. \cite{cat} utilizes horizontal and vertical rectangle window attention to expand the attention area and introduces the Axial-Shift operation for efficient window interactions. However, these methods generally restrict attention modeling to fixed windows with shift operations, limiting their ability to capture dynamic local attention. To strike a balance between capturing dependencies among relevant pixels and reducing computational complexity, we propose RouteWinFormer, a window-based transformer that can effectively capture middle-range attention and dynamically select the most relevant nearby windows in the cross and rectangle regions for attention aggregation.

\section{Method}
In this section, we first present the overall architecture of RouteWinFormer, followed by a detailed description of the Route-Windows Attention Module. Finally, we introduce the Multi-Scale Structure Regularization.

\subsection{Overall Architecture}
The workflow of RouteWinFormer is depicted in Fig.~\ref{arch}. It follows a U-shaped architecture with four scales. Given an input image $\mathbf{I} \in \mathbb{R}^{3 \times h \times w}$, a convolution is first applied to extract shallow features $\mathbf{F}_0 \in \mathbb{R}^{c \times h \times w}$, where $h$, $w$, and $c$ represent the height, width, and number of channels, respectively. The features $\mathbf{F}_0$ are then processed through the encoder-decoder structure. Each encoder and decoder scale contains several Transformer Blocks that extract context-enriched features $\mathbf{F}_i \in \mathbb{R}^{2^i \cdot c \times \frac{h}{2^i} \times \frac{w}{2^i}}$, where $i = 0, 1, 2, 3$. The encoder at each scale takes input from the preceding encoder level, downsampled using a  2×2 convolution with stride 2. Conversely, each decoder scale combines features from the corresponding encoder level and the upsampled output of the next-level decoder through an upsampling operator. This operator consists of a 1×1 convolution followed by PixelShuffle \cite{pixelshuffle}, which reduces the channel dimensions by half.

As illustrated in Fig. \ref{arch}(a), Block is primarily composed of an Image Restoration Block (IRBlock) and a Feed-Forward Network (FFN). The formulation of Block can be expressed as follows:
\begin{equation} \label{Block}
    \begin{aligned}
        \mathbf{X}&= \text{IRBlock}(\mathbf{F_{in}})+ \mathbf{F_{in}}\\
        \mathbf{Y} &= \text{FFN}(\mathbf{X}) + \mathbf{X}
    \end{aligned}
\end{equation}
where $\mathbf{F}_{\text{in}}$ represents the input features. IRBlock integrates two key branches: the Route-Windows Attention Module (RWAM) and the Shift-Windows Attention Module (SWAM) derived from Swin Transformer \cite{swin}. In addition, we simplify NAFBlock\cite{naf} as our FFN, which can be expressed as:
\begin{equation}\label{FFN}
    \begin{aligned}
        &\mathbf{X}=\text{Conv1×1}(\text{LN}(\mathbf{X})) \\
        &\mathbf{X}_t,\mathbf{X}_b=\text{Chunk}(\mathbf{X}) \\
        &\mathbf{X}=\text{GELU}(\text{DWConv}(\mathbf{X}_t))*\mathbf{X}_b \\
        &\mathbf{Y}=\text{Conv1×1}(\text{SCA}(\mathbf{X})*\mathbf{X})
    \end{aligned}
\end{equation}
where $\text{LN}$,$\text{Chunk}$,$\text{DWConv}$,$\text{SCA}$ denotes layer normalization, average split the channel of features, depth-wise convolution,  and simplified channel attention\cite{naf}, respectively.

\subsection{Route-Windows Attention Module}\label{RWAM}
To address the significant computational and memory overhead associated with directly leveraging long-range attention, we propose a novel Route-Windows Attention Module (RWAM), as illustrated in Fig.~\ref{arch}(b). RWAM incorporates a router that dynamically selects the most relevant nearby windows for attention aggregation based on regional similarity, efficiently extending the receptive field to a mid-range distance. To better adapt to diverse shapes of context and degradation patterns, we further introduce two types of regional similarity: cross-shaped and rectangle-shaped regions.

Specifically, given an input feature map $\mathbf{X}\in \mathbb{R}^{c\times h\times w}$, RWAM first divides it into non-overlap windows, resulting in a tensor $\textbf{X}\in \mathbb{R}^{s_h \times s_w \times c\times \frac{hw}{k^2}}$,where $s_h=\tfrac{h}{k}$ and $s_w=\tfrac{w}{k}$, and $k$ denotes the size of each window. The channels of the feature map are then evenly split into three chunks to derive $\mathbf{Q},\mathbf{K},\mathbf{V}\in \mathbb{R}^{s_h \times s_w \times d \times \frac{hw}{k^2}}$, where $d=\tfrac{c}{3}$.




\textbf{Router and Regional similarity.} The router begins by averaging the tensors $\textbf{Q}, \textbf{K} \in \mathbb{R}^{s_h \times s_w \times d \times \frac{hw}{k^2}}$ along the $\frac{hw}{k^2}$ dimension, yielding the window description vectors $\mathbf{Q}_r, \mathbf{K}_r \in \mathbb{R}^{s_h \times s_w \times d}$. The window description vector $\mathbf{K}_r$ is unfolded according to two types of regional similarity shape, resulting in $\mathbf{K}_r \in \mathbb{R}^{s_h \times s_w \times d \times r_n}$, where $r_n$ represents the number of candidate windows. Subsequently, router computes the regional similarity for each individual window to dynamically identify the relevant ones. The regional similarity matrix $\mathbf{H} \in \mathbb{R}^{s_h \times s_w \times r_n}$ is computed via matrix multiplication followed by a Softmax operation between $(\mathbf{Q}_r)^T \in \mathbb{R}^{s_h \times s_w \times 1 \times d}$ and $\mathbf{K}_r \in \mathbb{R}^{s_h \times s_w \times d \times r_n}$:
\begin{equation}\label{regional similarity}
    \begin{aligned}
        &\mathbf{H}=\text{Softmax}((\mathbf{Q}_r)^T\mathbf{K}_r) \\
    \end{aligned}
\end{equation}
To adapt to varying degradation and context patterns, we design two types of regional similarity: cross-shaped and neighbor-shaped regions, as illustrated in Fig.~\ref{arch}(c). These regional similarity calculation methods are alternated across blocks to accommodate different input modes. For each windows, $r_n$ candidate windows are identified, from which the relevant ones are selected based on the regional similarity matrix $\mathbf{H}$. The router employs the $\text{TopK}$ operation to extract the indices of the top-$k$ relevant windows while excluding irrelevant ones, thereby reducing computational and memory overhead. The selection process is formulated as:
\begin{equation}\label{select}
    \begin{aligned}
        &\mathbf{I}=\text{TopK}(\mathbf{H}) \\
    \end{aligned}
\end{equation}
Here, $\mathbf{I} \in \mathbb{R}^{s_h \times s_w \times r_i}$ contains the $r_i$ indices of the most relevant regions for each window, relative to the current center window. To convert these relative indices into their absolute counterparts within the original feature map, a mapping table(IndexTable) is constructed. This mapping is executed using the gather operation and is expressed as:
\begin{equation}\label{remap}
    \begin{aligned}
        \mathbf{J}=\text{gather}(\text{IndexTable}, \mathbf{I}) \\
    \end{aligned}
\end{equation}
where $\textbf{J}\in\mathbb{R}^{s_h \times s_w \times r_i}$ represents the absolute indices. Finally, the relative indices $\mathbf{I}$ are employed to retrieve relative position biases \cite{swin} from the relative position bias table $\hat{\mathbf{B}} \in \mathbb{R}^{r_n \times m \times (2k-1)^2}$, where $m$ denotes the number of attention heads. This process is formulated as:
\begin{equation}\label{bias}
    \begin{aligned}
        &\mathbf{B}_{1},\mathbf{B}_{2},\dots,\mathbf{B}_{r_i}=\text{gather}(\hat{\textbf{B}},\textbf{I}) \\
        &\mathbf{B}_g=\text{Concat}(\mathbf{B}_{1},\mathbf{B}_{2},\dots,\mathbf{B}_{r_i},\mathbf{B}_{c})
    \end{aligned}
\end{equation}
where $\mathbf{B}_{r_i} \in \mathbb{R}^{m \times k^2 \times k^2}$ represents the relative position biases for corresponding $r_i$ relevant windows of center window, $\mathbf{B}_{c}$ denotes the relative position bias for center windows, and $\mathbf{B}_g \in \mathbb{R}^{m \times k^2 \times (r_i+1) k^2}$ represents the combined relative position biases across all relevant windows.

\textbf{Middle-range Attention.} RWAM reformulates $\mathbf{Q}$, $\mathbf{K}$, $\mathbf{V}$ into $\mathbf{Q}_h$, $\mathbf{K}_h$, $\mathbf{V}_h \in \mathbb{R}^{s_h \times s_w \times m \times \frac{hw}{k^2} \times n}$ to enable multi-head attention, where $m$ denotes the number of attention heads and $n = \tfrac{d}{m} $ represents the dimension of each head. As illustrated in Fig.~\ref{arch}(b), RWAM utilizes the absolute indices $\textbf{J}$ to gather the corresponding windows for each individual center window and merges them as follows:
\begin{equation}\label{gather}
    \begin{aligned}
        &\mathbf{K}_1,\mathbf{K}_2,\dots,\mathbf{K}_{r_i}=\text{gather}(\textbf{K}_h,\textbf{J}) \\
        &\mathbf{V}_1,\mathbf{V}_2,\dots, \mathbf{V}_{r_i}=\text{gather}(\textbf{V}_h,\textbf{J}) \\
        & \textbf{K}_g=\text{Concat}(\mathbf{K}_1,\mathbf{K}_2,\dots ,\mathbf{K}_{r_i},\mathbf{K}_c)\\
        &\textbf{V}_g=\text{Concat}(\mathbf{V}_1,\mathbf{V}_2,\dots ,\mathbf{V}_{r_i},\mathbf{V}_c)
    \end{aligned}
\end{equation}
Here, $\textbf{K}_g,\textbf{V}_g\in \mathbb{R}^{s_h \times s_w \times m\times (r_i+1) \frac{hw}{k^2} \times n}$ are the merged key and value tensors, while $\mathbf{K}_c$ and $\mathbf{V}_c$ represent the key and value of the center window.  Then, RWAM applies middle-range attention for each window in $\textbf{Q}_h$:
\begin{equation}\label{attention}
    \begin{aligned}
        &\textbf{O}=\text{Softmax}(\frac{\textbf{Q}_h(\textbf{K}_g)^T}{\sqrt{d}}+\textbf{B}_g)\textbf{V}_g
    \end{aligned}
\end{equation}
where $\textbf{O}\in \mathbb{R}^{s_h \times s_w\times m \times \frac{hw}{k^2} \times n}$ represents the middle-range attention output. Finally, $\textbf{O}$ is reshaped back to the original input tensor shape, $\textbf{X}\in \mathbb{R}^{c\times h \times w}$. The computational complexity of the middle-range attention is reduced from $\mathcal{O}(h^2w^2c)$ to $\mathcal{O}(r_ik^2hwc)$, enabling the capture of pixel-level attention while maintaining a larger receptive field.

\begin{figure*}[!t]
    \includegraphics[width=\linewidth]{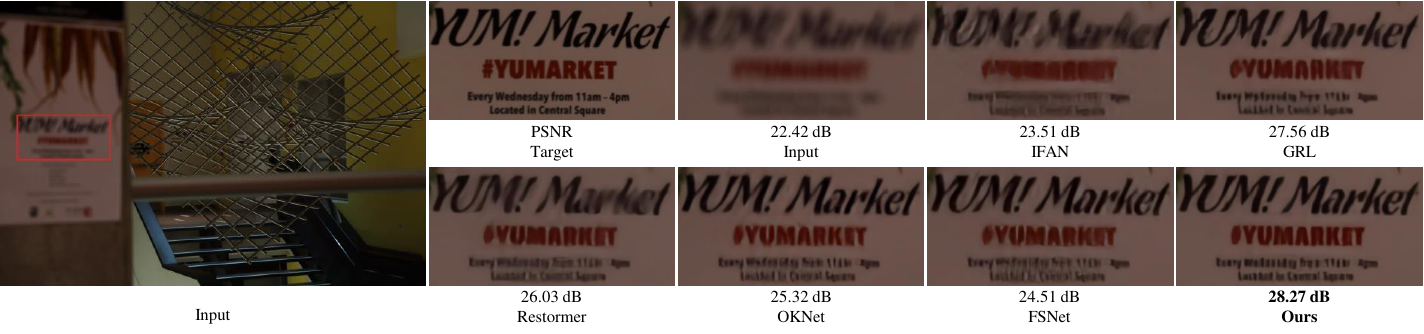}
    \caption{Visualize sample of image defocus deblurring on the DPDD dataset.}
    \label{DPDD_fig}
\end{figure*}

\begin{table*}[t]
\centering
\scalebox{0.94}{
\begin{tabular}{llcccccccccccc} 
\toprule
&& \multicolumn{4}{c}{\textbf{Indoor Scenes}} & \multicolumn{4}{c}{\textbf{Outdoor Scenes}} & \multicolumn{4}{c}{\textbf{Combined}} \\ \cmidrule(lr){3-6} \cmidrule(lr){7-10} \cmidrule(lr){11-14}
Method & Venue & PSNR & SSIM & MAE & LPIPS & PSNR & SSIM & MAE & LPIPS & PSNR & SSIM & MAE & LPIPS \\  \midrule
DMENet  & CVPR'19    & 25.50 & 0.788 & 0.038 & 0.298 & 21.43 & 0.644 & 0.063 & 0.397 & 23.41 & 0.714 & 0.051 & 0.349 \\
DPDNet  & ECCV'20    & 26.54 & 0.816 & 0.031 & 0.239 & 22.25 & 0.682 & 0.056 & 0.313 & 24.34 & 0.747 & 0.044 & 0.277 \\
KPAC    & ICCV'21    & 27.97 & 0.852 & 0.026 & 0.182 & 22.62 & 0.701 & 0.053 & 0.269 & 25.22 & 0.774 & 0.040 & 0.227 \\
IFAN    & CVPR'21    & 28.11 & 0.861 & 0.026 & 0.179 & 22.76 & 0.720 & 0.052 & 0.254 & 25.37 & 0.789 & 0.039 & 0.217 \\
Restormer  & CVPR'22 & 28.87 & 0.882 & 0.025 & 0.145 & 23.24 & 0.743 & 0.050 & 0.209 & 25.98 & 0.811 & 0.038 & 0.178 \\
FocalNet   & ICCV'23 & 29.10 & 0.876 & 0.024 & 0.173 & 23.41 & 0.743 & 0.049 & 0.246 & 26.18 & 0.808 & 0.037 & 0.210 \\
GRL       &  CVPR'23 & 29.06 & 0.886 & 0.024 & 0.139 & 23.45 & 0.761 & 0.049 & \underline{0.196} & 26.18 & 0.822 & 0.037 & 0.168 \\
OKNet     &  AAAI'24 & 28.99 & 0.877 & 0.024 & 0.169 & 23.51 & 0.751 & 0.049 & 0.241 & 26.18 & 0.812 & 0.037 & 0.206 \\
FSNet     &  TPAMI'24& 29.14 & 0.878 & 0.024 & 0.166 & 23.45 & 0.747 & 0.050 & 0.246 & 26.22 & 0.811 & 0.037 & 0.207 \\ 
CSNet     &  IJCAI'24& 29.20 & 0.881 & 0.023 & 0.147 & 23.45 & 0.752 & 0.049 & 0.206 & 26.25 & 0.815 & 0.037 & 0.178 \\
\midrule
\textbf{RWF-T} & \textbf{ours} & \underline{29.40} & \underline{0.889} & \underline{0.023} & \underline{0.138} & \underline{23.74} & \underline{0.765} & \underline{0.048} & 0.197 & \underline{26.49} & \underline{0.825} & \underline{0.036} & \underline{0.168} \\
\textbf{RWF-B} & \textbf{ours}   & \textbf{29.54} & \textbf{0.892} & \textbf{0.023} & \textbf{0.129} & \textbf{23.89} & \textbf{0.775} & \textbf{0.047} & \textbf{0.181} & \textbf{26.64} & \textbf{0.832} & \textbf{0.035} & \textbf{0.155} \\
\bottomrule
\end{tabular}
}
\caption{Image defocus deblurring comparisons on the DPDD (including 37 indoor and 39 outdoor scenes).}
\label{DPDD_tab}
\end{table*}

\subsection{Multi-Scale Structure Regularization}
Learning-based approaches often overfit to the training degradation patterns, rather than capturing the textures and structures of the image, resulting in suboptimal restoration performance \cite{MIM,ram}. To address this, we develop a Multi-Scale Structure Regularization (MSR) term in the loss function, which leverages the structure information from the clean image to generate additional learning regularization terms at the sub-scale level, helping to generalize the learning process. 

Specifically, at the $i$-th sub-scale, MSR uses the corresponding clean image as the ground truth. It then downsamples and upsamples the clean image to generate a texture-damaged version, which loses high-frequency details. Notably, the damaged image is only used for loss calculation and is not input to the network. MSR predicts the structure residual image $\mathbf{R}_i$, which captures the difference between the $i$-th ground truth and the damaged image. This residual image guides the sub-scales to focus on learning the textures and structures of the image. The $\mathcal{L}_{\text{MSR}}$ fomulation can be express as:
\begin{equation}\label{msr}
    \begin{aligned}
        \mathcal{L}_{\text{MSR}}=\sum^4_{i=2}\mathcal{L}(\mathbf{R}_i+\mathbf{G}_i^{\downarrow \uparrow},\mathbf{G}_i)
    \end{aligned}
\end{equation}
The term $\mathbf{G}_i^{\downarrow \uparrow}$ denotes the downsampling and the upsampling operations applied to the $i$-th ground-truth image. Furthermore, $\mathbf{R}_i$ represents the predicted residual image at $i$-th scale. Consequently, the loss function with $\mathcal{L}_{\text{MSR}}$ can be expressed as follows:
\begin{equation}\label{loss_all}
    \begin{aligned}
        \mathcal{L}_{all}=\mathcal{L}(\hat{\mathbf{I}},\mathbf{G})+\lambda \mathcal{L}_{\text{MSR}}
    \end{aligned}
\end{equation}
where $\hat{\mathbf{I}},\mathbf{G},\lambda$ represent the restored image, the ground-truth image and loss weight, respectively. The $\hat{\mathbf{I}}$ is obtained using the formula: $\hat{\mathbf{I}}=\mathbf{I}_{in}+\mathbf{I}_{rs}$, where the $\mathbf{I}_{in}$ is the input image and the $I_{rs}$ is residual image. In addition, we incorporate a frequency domain loss to further enhance image restoration performance. The $\mathcal{L}$ can be defined as follows:
\begin{equation}\label{loss_normal}
    \begin{aligned}
        \mathcal{L}(\mathbf{X},\mathbf{Y})=\mathcal{L}_1(\mathbf{X},\mathbf{Y})+\alpha \mathcal{L}_{\text{FFT}}(\mathbf{X},\mathbf{Y})
    \end{aligned}
\end{equation}
where $\alpha$ are empirically default set to $0.1$, respectively. The $\mathcal{L}_1$ loss can be formulated as follows:
\begin{equation}\label{loss_l1}
    \begin{aligned}
        \mathcal{L}_1=\vert\vert \mathbf{X}-\mathbf{Y} \vert\vert_1
    \end{aligned}
\end{equation}
Then, $\mathcal{L}_{\text{FFT}}$ represents the frequency domain $\mathcal{L}_1$ loss:
\begin{equation}\label{loss_fft}
    \begin{aligned}
        \mathcal{L}_{\text{FFT}}=\vert\vert \text{FFT}(\mathbf{X})-\text{FFT}(\mathbf{Y})\vert\vert_1
    \end{aligned}
\end{equation}
where $\text{FFT}$ represents fast Fourier
transform.

\begin{table}[t]
\centering
\scalebox{0.92}{
\begin{tabular}{cccccc}
\toprule
Model & ${[}L_1-L_4{]}$   & C  & Params(M) & Flops(G) & Time/s \\ \midrule
RWF-T & ${[}2,4,6,8{]}$ & 32 & 11.15     & 25.29    &  0.0197      \\
RWF-S & ${[}2,4,4,8{]}$ & 48 & 23.23     & 47.77    &  0.0201      \\
RWF-B & ${[}2,4,6,8{]}$ & 64 & 43.06     & 89.79    &  0.0268      \\ \bottomrule
\end{tabular}
}
\caption{Hyperparameters of RouteWinFormer. FLOPs are measured for a single 256×256 image, while inference time is averaged with 1280 images.}
\label{config}
\end{table}

\section{Experiments}\label{Experiments}
In this section, we evaluate the performance of RouteWinFormer on 9 widely used datasets covering four tasks: image defocus deblurring, image desnowing, image dehazing, and image deraining. In the tables, the highest quality scores for each evaluated method are bolded, with the second-best results underlined.

\subsection{Implementation Details}
To balance performance and efficiency, we propose three versions of RouteWinFormer (RWF): Tiny (RWF-T), Small (RWF-S), and Base (RWF-B). The different hyperparameters are summarized in Table \ref{config}. Multi-head settings for RWAM and SWAM follow $[1, 2, 4, 8]$, with the default number of select windows $r_i$ for RWAM set to $[1, 1, 1, 1]$. All models are optimized using the AdamW optimizer ($\beta_1=0.9$, $\beta_2=0.999$) for 300K iterations. The learning rate starts at $1 \times 10^{-3}$ and decays to $1 \times 10^{-7}$ via cosine annealing \cite{sche}. Data augmentation, including horizontal and vertical flips, is applied during training on 256×256 patch images. To mitigate performance degradation due to patch-based training versus full-image testing, we employ Test-Time Local Coordinate (TLC) \cite{tlc}. All experiments are performed on an NVIDIA GeForce RTX 3090 GPU.

\subsection{Experiments Results}
\textbf{Image Defocus Deblurring.} The evaluation of single-image defocus deblurring is conducted on the DPDD dataset \cite{dpdd}, with results presented in Table~\ref{DPDD_tab}. RouteWinFormer significantly outperforms state-of-the-art methods across all scene categories. Compared to FSNet~\cite{FSNet}, it achieves PSNR improvements of 0.42 dB for combined scenes, 0.40 dB for indoor scenes, and 0.44 dB for outdoor scenes. In addition, it surpasses CSNet~\cite{hybrid} and OKNet~\cite{omni} by 0.39 dB and 0.46 dB, respectively, on combined scenes. Notably, RouteWinFormer achieves a 0.46 dB PSNR gain over the global attention Transformer GRL \cite{grl}. Qualitative results in Fig.~\ref{DPDD_fig} emphasize that RouteWinFormer is more effective in recovering sharper details than the compared algorithms.

\begin{figure*}[!t]
    \includegraphics[width=\linewidth]{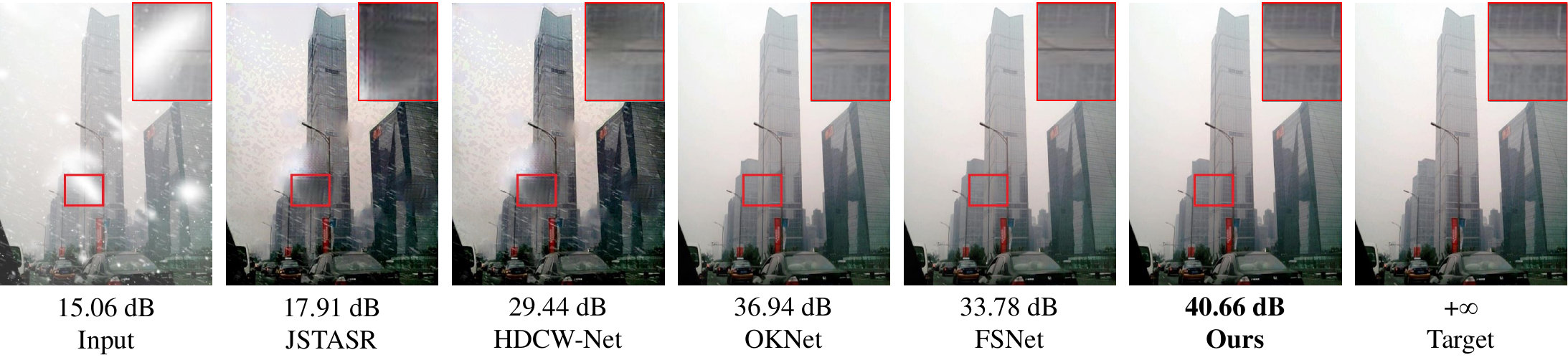}
    \caption{Visualize sample of image desnowing on the CSD dataset.}
    \label{csd_fig}
\end{figure*}

\begin{figure*}[t]
    \includegraphics[width=\linewidth]{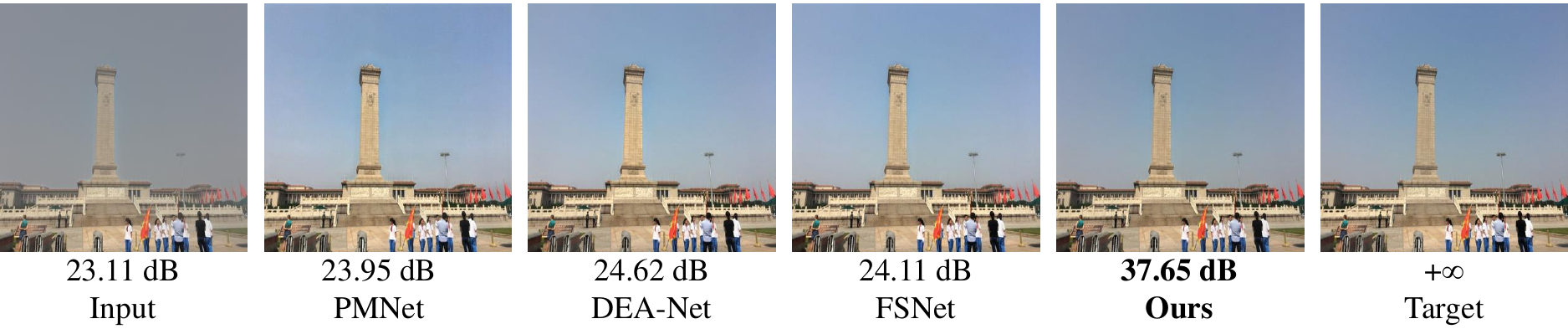}
    \caption{Visualize sample of image dehazing on the Haze4K dataset.}
    \label{haze4k_fig}
\end{figure*}

\begin{table}[t]
\centering
\begin{tabular}{lcccc}
\toprule
& \multicolumn{2}{c}{CSD} & \multicolumn{2}{c}{SRRS}\\ 
\cmidrule(lr){2-3} \cmidrule(lr){4-5}
Method & PSNR & SSIM & PSNR & SSIM \\ 
\midrule
DesnowNet    & 20.13 & 0.81 & 20.38 & 0.84 \\
CycleGAN     & 20.98 & 0.80 & 20.21 & 0.74\\
JSTASR          & 27.96 & 0.88 & 25.82 & 0.89 \\
HDCW-Net         & 29.06 & 0.91 & 27.78 & 0.92\\
SMGARN       & 31.93 & 0.95 & 29.14 & 0.94 \\
TransWeather & 31.76 & 0.93 & 28.29 & 0.92 \\
MSP-Former & 33.75 & 0.96 & 30.76 & 0.95\\
FocalNet          & 37.18 & 0.99 & 31.34 & 0.98\\
OKNet            & 37.99 & 0.99 & 31.70 & 0.98 \\
FSNet  & \underline{38.37} & \underline{0.99} & \underline{32.33} & \underline{0.98} \\
\midrule
\textbf{RWF-S}  & \textbf{40.08} & \textbf{0.99} & \textbf{32.37} & \textbf{0.98}\\ 
\bottomrule
\end{tabular}
\caption{Image desnowing comparison results on the CSD and SRRS datasets.}
\label{desnow_tab}
\end{table}

\begin{table}[t]
\centering
\begin{tabular}{lcccc}
\toprule
& \multicolumn{2}{c}{Haze4K}  & \multicolumn{2}{c}{NH-HAZE}      \\ \cmidrule(lr){2-3} \cmidrule(lr){4-5}
Method     & PSNR           & SSIM          & PSNR           & SSIM            \\ 
\midrule
DehazeNet   & 19.12          & 0.84        & 16.62          & 0.52                 \\
AOD-Net     & 17.15          & 0.83        & 15.40          & 0.57         \\
GridDehazeNet  & 23.29          & 0.93     & 13.80          & 0.54          \\
KDDN        & -          & -    & 17.39          & 0.59          \\
MSBDN       & 22.99          & 0.85        & 19.23          & 0.71          \\
FFA-Net     & 26.96          & 0.95        & 19.87          & 0.69            \\
PMNet       & 33.49          & 0.98        & 20.42          & 0.73          \\
FSNet       & 34.12          & 0.99        & 20.55          & \textbf{0.81}\\ 
DEA-Net       & 34.25          & 0.99        & -          & - \\ 
ConvIR       & \underline{34.50} & \underline{0.99} & \underline{20.66} & \underline{0.80}\\ 
\midrule
\textbf{RWF-S}        & \textbf{35.08} & \textbf{0.99} & \textbf{21.49} & 0.69  \\ 
\bottomrule
\end{tabular}
\caption{Image dehazing results on the Haze4K and NH-HAZE datasets.}
\label{dehaze_tab}
\end{table}

\textbf{Image Desnowing.} Following prior works \cite{FSNet}, we evaluate RouteWinFormer on the CSD~\cite{csd} and SRRS~\cite{srrs} datasets using PSNR and SSIM metrics. As shown in Table~\ref{desnow_tab}, RouteWinFormer achieves a significant 1.71 dB improvement over FSNet~\cite{FSNet} on the CSD dataset. On SRRS, it outperforms OKNet~\cite{omni} by over 0.6 dB. Compared to MSP-Former~\cite{msp-former}, RouteWinFormer gains 6.33 dB on CSD and 1.61 dB on SRRS. Visual results in Fig.~\ref{csd_fig} further demonstrate its effectiveness in removing snowflakes while preserving fine details.

\textbf{Image Dehazing.} The image dehazing comparison is performed on the Haze4K \cite{haze4k} and NH-HAZE \cite{nhhaze} dataset. As shown in Table \ref{dehaze_tab}, RouteWinFormer consistently outperforms recent methods across both datasets. On Haze4K, it improves PSNR by 0.58 dB over ConvIR \cite{ConvIR} and 0.96 dB over FSNet \cite{FSNet}. On NH-HAZE, RouteWinFormer achieves a 0.83 dB gain over ConvIR\cite{ConvIR} and 0.9 dB over OKNet \cite{omni}. Visual results in Fig.~\ref{haze4k_fig} further highlight our method’s ability to produce high-quality dehazed images.

\begin{figure*}[!t]
    \includegraphics[width=\linewidth]{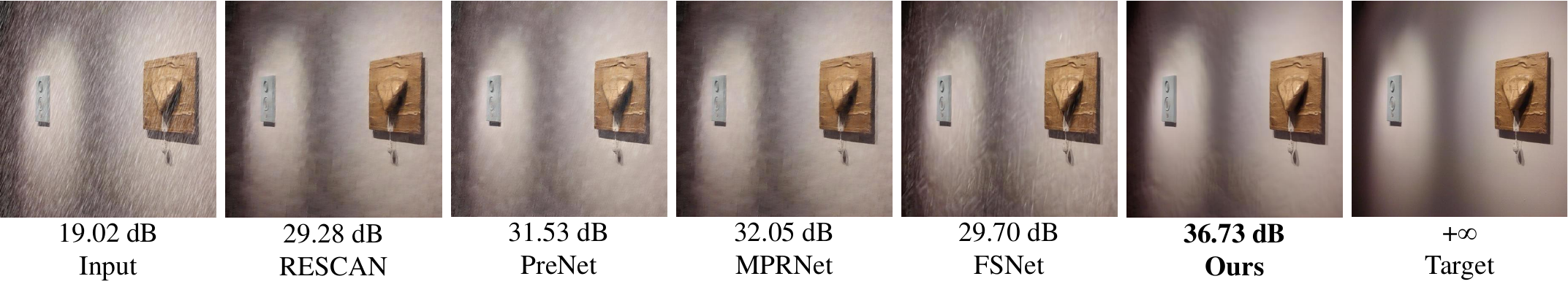}
    \caption{Visualize sample of image deraining on the Test1200 dataset.}
    \label{derain_fig}
\end{figure*}

\begin{figure*}[t]
    \centering
    \includegraphics[width=\linewidth]{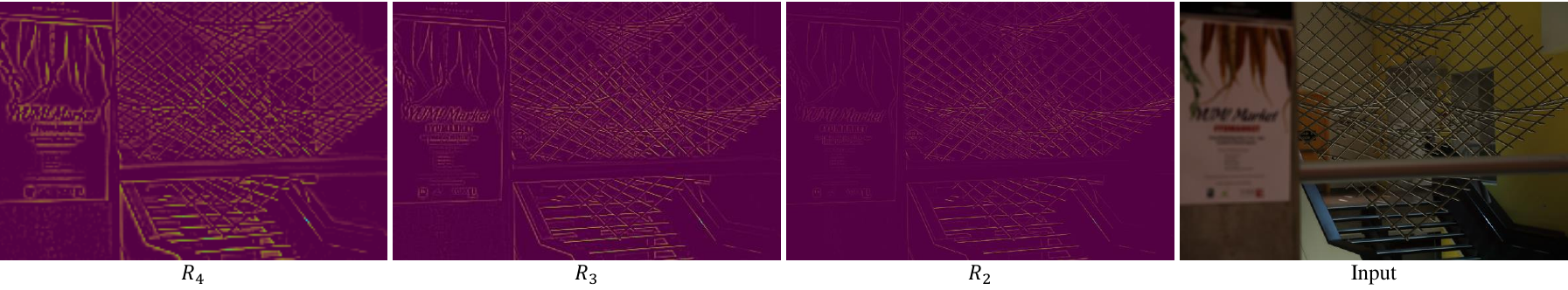}
    \caption{Visualize sample of MSR on DPDD dataset.}
    \label{msr_fig}
\end{figure*}

\begin{table}[t]
\centering
\scalebox{0.85}{
\begin{tabular}{lcccccc}
\toprule
& \multicolumn{2}{c}{Rain100H} & \multicolumn{2}{c}{Rain100L} & \multicolumn{2}{c}{Test1200} \\ 
\cmidrule(lr){2-3} \cmidrule(lr){4-5} \cmidrule(lr){6-7}
Method & PSNR & SSIM & PSNR & SSIM & PSNR & SSIM \\ 
\midrule
DerainNet   & 14.92 & 0.592 & 27.03 & 0.884 & 23.38 & 0.835 \\
UMRL    & 26.01 & 0.832 & 29.18 & 0.923 & 30.55 & 0.910 \\
RESCAN    & 26.36 & 0.786 & 29.80 & 0.881 & 30.51 & 0.882 \\
PreNet    & 26.77 & 0.858 & 32.44 & 0.950 & 31.36 & 0.911 \\
MSPFN    & 28.66 & 0.860 & 32.40 & 0.933 & 32.39 & 0.916 \\
MPRNet    & 30.41 & 0.890 & 36.40 & 0.965 & 32.91 & 0.916 \\
HINet    & 30.65 & 0.894 & 37.28 & 0.970 & 33.05 & \underline{0.919} \\
FSNet          & \textbf{31.77} & \textbf{0.906} & \underline{38.00} & \underline{0.972} & \underline{33.08} & 0.916 \\
ConvIR         & 31.09 & 0.903 & 37.69     & 0.970 & 31.19     & 0.894     \\ \midrule
\textbf{RWF-S}   & \underline{31.49} & \underline{0.904} & \textbf{38.92} & \textbf{0.977} & \textbf{33.69} & \textbf{0.927} \\ 
\bottomrule
\end{tabular}
}
\caption{Image Deraining results on the Rain100H, Rain100L and Test1200 datasets.}
\label{deraining_tab}
\end{table}

\begin{table}[t]
\centering
\scalebox{0.84}{
\begin{tabular}{lccccc}
\toprule
Dataset & Method & PSNR & SSIM & Params (M) & Flops (G) \\ \midrule
\multirow{2}{*}{CSD}  & Baseline-S & 39.58 & 0.990 & 22.86 & 45.36 \\
                      & RWF-S    & 39.89 & 0.990 & 23.23 & 47.77 \\ \midrule
\multirow{2}{*}{DPDD} & Baseline-T & 26.08 & 0.809 & 10.75 & 23.55 \\
                      & RWF-T    & 26.24 & 0.817 & 11.15 & 25.29 \\ 
\bottomrule
\end{tabular}
}
\caption{Ablation studies for RWAM on the CSD and DPDD Combined Scenes.}
\label{ab1}
\end{table}

\begin{table}[t]
\centering
\scalebox{0.87}{
\begin{tabular}{ccccccc}
\toprule
& \multicolumn{2}{c}{CSD} & \multicolumn{2}{c}{$\text{DPDD}_I$} & \multicolumn{2}{c}{$\text{DPDD}_C$} \\ \cmidrule(lr){2-3} \cmidrule(lr){4-5} \cmidrule(lr){6-7}
        & PSNR       & SSIM       & PSNR           & SSIM        & PSNR            & SSIM             \\ \midrule
w\_MSR  & 39.03      & 0.9883     & 28.96          & 0.8783      & 26.08           & 0.8094           \\
wo\_MSR & 38.86      & 0.9879     & 28.51          & 0.8714      & 25.82           & 0.8042           \\ \bottomrule
\end{tabular}
}
\caption{Ablation study for MSR. $\text{DPDD}_I$ and $\text{DPDD}_C$ represent the Indoor Scenes and Combined Scenes in the DPDD dataset.}
\label{ab2}
\end{table}

\textbf{Image DeRaining.} To evaluate performance on image deraining tasks, we calculate the PSNR and SSIM scores using the Y channel in the YCbCr color space, following~\cite{FSNet}. Quantitative results in Table~\ref{deraining_tab} show a 0.61 dB PSNR improvement on the Test1200 dataset over FSNet~\cite{FSNet}. Moreover, RouteWinFormer outperforms ConvIR~\cite{ConvIR} on the Rain100H and Rain100L datasets, achieving PSNR scores of 31.49 dB and 38.92 dB, respectively, compared to 31.09 dB and 37.69 dB. Visual results in Fig.~\ref{derain_fig} demonstrate that our method effectively removes rain while preserving intricate details and producing high-quality derained images.

\subsection{Ablation Study}
In this subsection, we present ablation experiments conducted on the CSD and DPDD datasets. The baseline model consists only of the SWAM module with FFN in the Block.

\textbf{Effect of RWAM.} We conduct an ablation study on the CSD and DPDD Combined Scenes to evaluate the effect of RWAM. As shown in Table~\ref{ab1}, RWAM achieves a 0.31 dB PSNR gain with a small increase of 0.37 M parameters and 2.41 GFlops on CSD. Compared to the RWF-T model on DPDD, RWAM improves PSNR by 0.16 dB with 11.15 M parameters and 25.29 GFlops. These results demonstrate that RWAM effectively extends the receptive field and enhances image restoration.


\textbf{Effect of MSR.} To demonstrate that MSR is model-agnostic, we conduct experiments on the baseline-T, as shown in Table \ref{ab2}. When coupled with MSR, the baseline achieves a 0.17 dB increase on the CSD dataset, 0.45 dB on the Indoor Scenes of the DPDD dataset, and 0.23 dB on the Combined Scenes of the DPDD dataset. Visual results in Fig.~\ref{msr_fig} further highlight effectiveness of MSR in capturing textures.

\section{Conclusion}
In this paper, we propose RouteWinFormer, a window-based Transformer for image restoration that models middle-range context. It incorporates the Route-Windows Attention Module, dynamically selecting nearby windows based on regional similarity for efficient attention aggregation. Additionally, Multi-Scale Structure Regularization helps the sub-scales of the U-shaped network focus on structural information, while the original-scale learns degradation patterns.

\appendix

\section*{Ethical Statement}

There are no ethical issues.



\bibliographystyle{named}
\bibliography{ijcai25}
\end{document}